\documentclass{article}
\usepackage{PRIMEarxiv}

\usepackage{hyperref}
\usepackage{color}
\usepackage{pifont}
\usepackage{paralist}
\usepackage{subfigure}
\usepackage{microtype}
\usepackage{fancyhdr}       
\usepackage{graphicx}       
\usepackage{xcolor}
\usepackage{multirow}

\usepackage{cite}
\usepackage{url}
\usepackage{hyperref}
\usepackage{datetime}
\usepackage{booktabs}

\usepackage{array} 
\usepackage{threeparttable}
\usepackage{colortbl, xcolor}

\usepackage[most]{tcolorbox}
\usepackage{varwidth}
\newtcolorbox{resbox}[2][]{enhanced,
before skip=2mm, after skip=2mm,
colback=cyan!6!white, colframe=cyan!40!blue!70, boxrule=0.25mm,
attach boxed title to top left={xshift=0.4cm, yshift*=0.7mm-\tcboxedtitleheight},
varwidth boxed title*=-3cm,
boxed title style={frame code={
    \path[fill=cyan!30!blue!60!black]
      ([yshift=-1mm,xshift=-1mm]frame.north west)
        arc[start angle=0,end angle=180,radius=1mm]
      ([yshift=-1mm,xshift=1mm]frame.north east)
        arc[start angle=180,end angle=0,radius=1mm];
    \path[left color=cyan!50!blue!70!black, right color=cyan!50!blue!70!black,
      middle color=cyan!60!blue!80!black]
      ([xshift=-2mm]frame.north west) -- ([xshift=2mm]frame.north east)
      [rounded corners=1mm]-- ([xshift=1mm,yshift=-1mm]frame.north east)
      -- (frame.south east) -- (frame.south west)
      -- ([xshift=-1mm,yshift=-1mm]frame.north west)
      [sharp corners]-- cycle;
    }, interior engine=empty,
  },
fonttitle=\bfseries\color{white}, title={#2}, #1}

\usepackage{cleveref}
 
\definecolor{headerblue}{RGB}{115, 178, 228}
\definecolor{columnblue}{RGB}{208, 233, 250}

\newcommand{\approach}{{\mbox{RoboULM}}}
\newcommand{\taxonomy}{{\mbox{UncerTax}}}

\title{
Human-in-the-Loop Uncertainty Analysis in Self-Adaptive Robots Using LLMs
}

\author{
  Hassan Sartaj \\
  Simula Research Laboratory \\
  Oslo, Norway\\
  \texttt{hassan@simula.no} \\
   \And
  Jalil Boudjadar \\
  Aarhus University \\
  Aarhus, Denmark\\
  \texttt{jalil@ece.au.dk} \\
  \And
  Mirgita Frasheri \\
  Aarhus University \\
  Aarhus, Denmark\\
  \texttt{mirgita.frasheri@ece.au.dk} \\
  \And
  Shaukat Ali \\
  Simula Research Laboratory \\
  Oslo, Norway\\
  \texttt{shaukat@simula.no} \\
  \And
  Peter Gorm Larsen \\
  Aarhus University \\
  Aarhus, Denmark\\
  \texttt{pgl@ece.au.dk} \\
}

\begin{document}
\maketitle

\begin{abstract}
Self-adaptive robots operate in dynamic, unpredictable environments where unaddressed uncertainties can lead to safety violations and operational failures. 
However, systematically identifying and analyzing these uncertainties, including their sources, impacts, and mitigation strategies, remains a significant challenge given the inherent complexity of real-world environments, dynamic robotic behavior, and the rapid evolution of robotic technologies. 
To address this, we introduce \approach{}, a human-in-the-loop methodology and tool that supports practitioners in systematically exploring uncertainties at the design stage using large language models (LLMs). 
Moreover, we present an uncertainty taxonomy that provides a detailed catalog of uncertainties in self-adaptive robots. 
We evaluated \approach{} with 16 practitioners from four industrial use cases. 
The results show that \approach{} was perceived as both useful and easy to understand, with the participants particularly valuing structured prompting and iterative refinement support. 
These findings demonstrate the potential of \approach{} as a viable solution for systematic uncertainty analysis in complex robots.
\end{abstract}

\keywords{Self-Adaptive Systems \and Robotics \and Uncertainty \and Large Language Models, \and Taxonomy}

\section{Introduction}
Self-adaptive robots (SARs) are subject to various uncertainties throughout the software development lifecycle and real-world operations. 
Early identification of uncertainties, their sources, and potential impacts at design time is considerably more cost-effective than addressing them post-deployment, where they can lead to costly redesigns or unsafe operational behavior~\cite{isaku2025oodisar}. 
In practice, practitioners and researchers often rely on intuition, prior experience, taxonomies, and quantification techniques to identify and manage uncertainties~\cite{sartaj2025identifying,takemura2024uncertainty}. 
However, these typically uncover only a limited subset of uncertainties and struggle to keep up with evolving robotics technologies and the integration of LLMs. 
Given the growing complexity of SARs, achieving systematic and rigorous uncertainty analysis remains a significant unresolved challenge. 

Recently, LLMs have demonstrated exceptional capabilities in many domains, including robotics~\cite{firoozi2025foundation}. 
Their advanced reasoning, contextual understanding, and knowledge synthesis abilities make them promising candidates for supporting complex engineering tasks. 
Motivated by this potential, our previous work~\cite{sartaj2025identifying} investigated the feasibility of using LLMs to identify uncertainties in SARs. 
The findings indicated that LLMs can effectively identify uncertainty while also revealing previously overlooked uncertainties and interesting scenarios. 
When combined with expert guidance and domain-specific reasoning, LLMs can systematically support exploring uncertainties in a more structured and comprehensive manner.

In this article, we propose a methodology for systematic uncertainty analysis in 
SARs comprising three key novel elements.
\begin{inparaenum}[(i)]
    \item An uncertainty taxonomy (\taxonomy{}) for SARs that provides a structured categorization of uncertainties, derived from four industrial case studies.
    \item A human-in-the-loop approach and tool (\approach{}) for systematic exploration of uncertainties in robots at design time using their requirements, applying \taxonomy{} to guide LLM reasoning. 
    \item Advanced prompting strategies and refinement methods for robotic engineers to iteratively refine LLM-generated responses through: \textit{ranking-based refinement} for rating response segments, \textit{taxonomy-guided refinement} for directing LLM reasoning using \taxonomy{}, and \textit{example-driven refinement} for instructing the LLM with concrete real-world examples or scenarios.
\end{inparaenum}

To evaluate \approach{}, we conducted a study with 16 practitioners of various roles and experience, from four real-world robotic use cases: Autonomous Mobile Robot (AMR), Industrial Disassembly Robot (IDR), Collaborative Manufacturing Robot (CMR), and Autonomous Vessel (AV). 
The results demonstrate that \approach{} received strong ratings for both utility and ease of understanding. 
In particular, features such as structured prompting and example-driven refinement were consistently identified as the most preferred across all cases, and participants highlighted the tool’s support for iterative refinement of LLM responses as the most helpful and efficient aspect. Overall, these findings indicate the practicality of \approach{} for systematic uncertainty exploration in complex robots.

\section{Real-World Robots}\label{sec:context} 
The Horizon Europe (HEU) project (RoboSAPIENS) develops methods and tools to engineer software for SARs, enabling them to handle uncertain and unknown situations autonomously by triggering appropriate adaptations. The project extends the traditional MAPE-K~\cite{kephart2003vision} feedback loop architecture with self-adaptation by introducing a new phase called \textit{Legitimate}, resulting in the MAPLE-K loop. The purpose of this additional phase is to verify and validate the proposed adaptations against relevant safety requirements before they are applied to the robots. Since uncertainty can arise from the SARs themselves (e.g., embodied deep learning components) as well as from the unpredictable environments in which they operate, it is important to develop methods to systematically identify and manage uncertainties in these systems. 
As part of RoboSAPIENS, we have access to four SARs, which are described below.

\subsection{Autonomous Mobile Robot}
The AMR features precise maneuverability in confined spaces and support for autonomous navigation, obstacle avoidance, and path planning. 
This use case focuses on safe and intelligent fleet navigation involving operating AMRs as a coordinated fleet in dynamic environments, using a global planner to determine optimal routes on predefined maps and a local planner that continuously adapts trajectories to moving obstacles and changing conditions.

\subsection{Industrial Disassembly Robot} 
The IDR is based on a Franka Emika Panda 7-DOF manipulator that supports handling tools, unsnapping tools, and an electric screwdriver for fine manipulation tasks on laptops. 
This use case focuses on trustworthiness in adaptive laptop refurbishment to automate key disassembly steps in laptop remanufacturing, such as screen and battery replacement, by learning force-based manipulation primitives from human expert demonstrations. 

\subsection{Collaborative Manufacturing Robot} 
The CMR operates in a modern factory setting alongside humans, other machines, and automated guided vehicles, where humans and machines frequently share space and tasks in a highly dynamic shop floor.
The CMR use case focuses on developing dynamic risk models that enable the robot to adjust its behavior and safety measures in real time. 

\subsection{Autonomous Vessel} 
The AV represents R/V Gunnerus, a ship equipped with sensors such as GPS and IMU that can provide detailed motion data during operation. 
The AV use case focuses on developing self-adaptive motion prediction models that handle structural changes and environmental disturbances by combining dynamic ship models with data-driven approaches and co-simulation of hull, propulsion, environment, and control.

\begin{table*}[htbp]
    \centering
    \scriptsize
    \setlength{\tabcolsep}{5pt}
    \renewcommand{\arraystretch}{1.5}
    \caption{\taxonomy{} Part-1: Uncertainty Identification Methods in SARs}
    \label{tab:taxonomy_identification}
    \begin{threeparttable}
    \begin{tabular}{|>{\bfseries\raggedright\arraybackslash\columncolor{columnblue}}p{2.7cm}|c|c|c|c|c|c|c|c|c|c|c|c|}
    \hline
    \rowcolor{headerblue}
    \textbf{Identification} &
    \textbf{Nature} &
    \textbf{Type} &
    \textbf{Stage} &
    \textbf{Temporal} &
    \textbf{Occur.} &
    \textbf{Adapt.} &
    \textbf{Scope} &
    \textbf{Risk} &
    \textbf{Affect} &
    \textbf{Propag.} &
    \textbf{Data} &
    \textbf{Ethical} \\ \hline

        Hardware specifications &
        St, De & Epis & Des & Long-term & HW & Internal & L, C & Low & S, R & Isolated & Pre & Tran\\ \hline

        Assembling hardware parts &
        St, De & Epis & Dev & Short-term & HW, Env & Internal & L, C & Mod & S, P & Cascading & Amb & Bias\\ \hline

        Operations/ field testing &
        Dy, So & Alea & Ops & Short-term & Env, SW & External & G, S & High & S, R & Cascading & Noi & Fair \\ \hline

        Analyzing behavioral deviations &
        Dy, So & Epis & Tes & Long-term & SW & Internal & G, S & High & P, A & Cascading & Inc & Trust \\ \hline

        Formal modeling with nondeterminism &
        Dy, So & Epis & Des & Long-term & SW, Env & Internal & G, S & High & P, A & Isolated & Pre & Tran\\ \hline

        Intuition &
        Dy, So & Epis & Des & Short-term & Human & Internal & L, C & Mod & S, R & Isolated & Amb & Bias\\ \hline

        Proof of concept demonstration &
        St, De & Epis & Tes & Short-term & HW, Env & Internal & L, C & Low & S, P & Isolated & Pre & Fair \\ \hline

        Component variations &
        St, De & Alea & Ops & Short-term & HW & External & L, C & Mod & S, R & Cascading & Noi & Tran\\ \hline

        Sensor data analysis &
        Dy, So & Epis & Dev & Short-term & Env, HW & Internal & L, C & High & S, R & Cascading & Noi & Trust \\ \hline

    \end{tabular}
    \begin{tablenotes}
    \scriptsize
    \item \textbf{Nature:} St = Static; De = Deterministic; Dy = Dynamic; So = Stochastic. \quad
    \textbf{Type:} Epis = Epistemic; Alea = Aleatoric. \quad
    \textbf{Stage:} Des = Design; Dev = Development; Tes = Testing; Ops = Operational. \quad
    \textbf{Occurrence:} HW = Hardware; Env = Environmental; SW = Software. \quad
    \textbf{Scope:} L = Local; C = Component; G = Global; S = System. \quad
    \textbf{Risk:} Mod = Moderate. \quad
    \textbf{Affect:} S = Safety; R = Reliability; P = Performance; A = Adaptability. \quad
    \textbf{Data:} Pre = Precise; Amb = Ambiguous; Noi = Noisy; Inc = Incomplete. \quad
    \textbf{Ethical:} Tran = Transparency; Fair = Fairness.
    \end{tablenotes}
    \end{threeparttable}
\end{table*}

\begin{figure*}[!t]
\centering
\includegraphics[width=\textwidth]{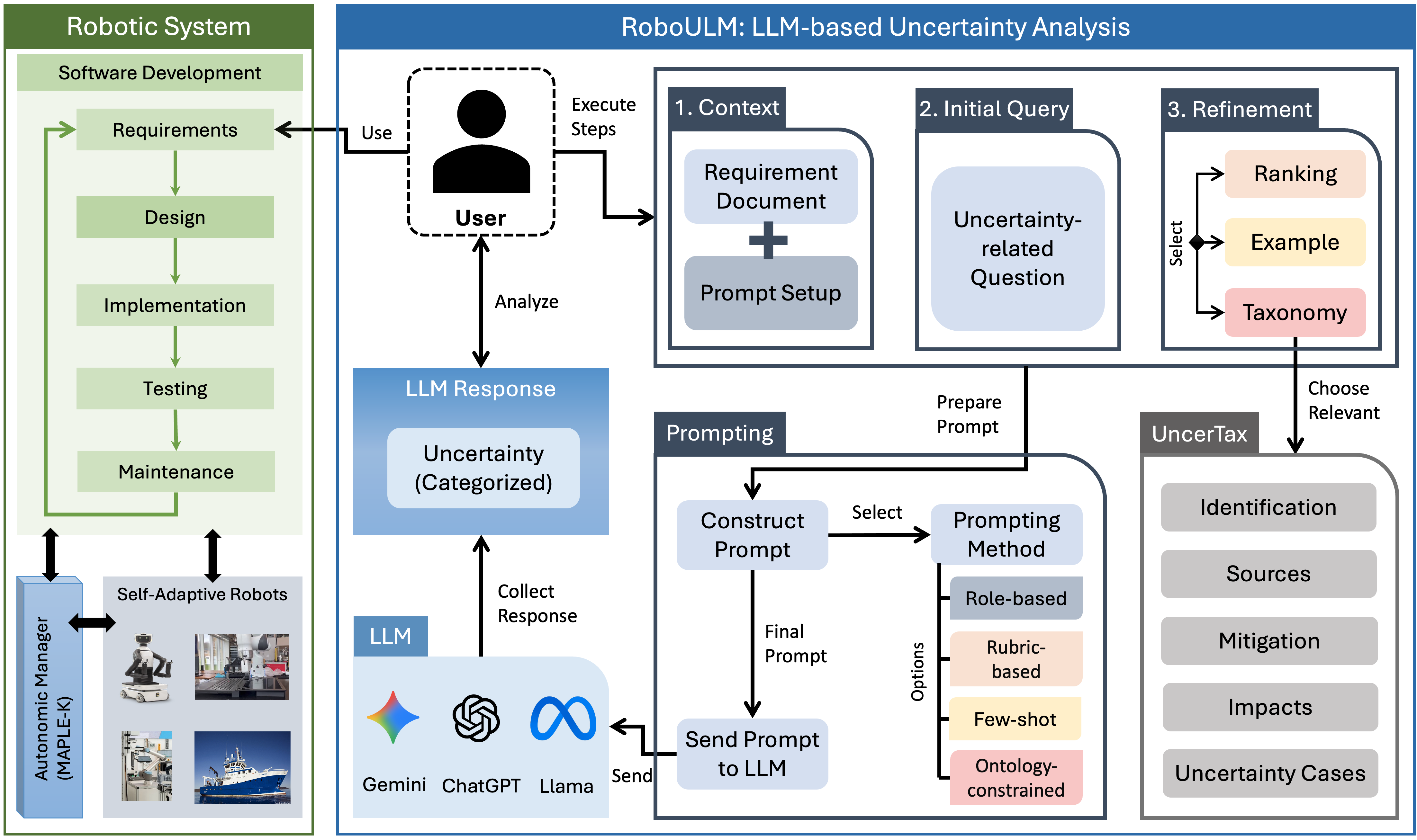}
\caption{Overview of \approach{} and its integration into robotic development process, illustrating human-in-the-loop design-time uncertainty analysis with large language models (LLMs) based on system requirements.}
\label{fig_approach}
\end{figure*}

\section{Methodology}\label{sec:approach}
As shown in~\Cref{fig_approach}, the proposed methodology relies on an uncertainty taxonomy, a cognitive exploration using LLM to identify uncertainties following user prompts, and refinement processes to direct and instruct the LLM inference.

\subsection{Uncertainty Taxonomy}\label{ssec:taxonomy}
We developed an uncertainty taxonomy (\taxonomy{}) for SARs, derived from survey data collected from industry practitioners working with four different robots and from LLM-generated outputs reviewed and validated by practitioners, both of which are reported in the same study~\cite{sartaj2025identifying}. 
\taxonomy{} is organized into five main aspects: commonly used \textit{identification} methods, \textit{sources} of uncertainty, their \textit{impacts}, \textit{mitigation} techniques, and real-world \textit{uncertainty cases}.
Within these aspects, uncertainty is characterized along 12 dimensions: nature (origin of uncertainty), type (aleatory or epistemic), stage (lifecycle phase), temporal (duration and persistence), occurrence, adaptation source, scope (local/global), risk/severity, affect, propagation (across components), data characteristics, and ethical implications. 
These dimensions were systematically identified from existing taxonomies and surveys on uncertainty in robotics and self-adaptive systems~\cite{Harriet24,ramirez2012taxonomy,hezavehi2021uncertainty}.

\Cref{tab:taxonomy_identification} presents the \taxonomy{} part on \textit{uncertainty identification methods} commonly used for SARs. 
To illustrate, consider hardware specifications: at the design stage, engineers inspect datasheets to identify measurement uncertainties expressed as tolerance ranges for long-term impact analysis. 
For example, a sensor rated at ±0.5mm represents a static and deterministic form of uncertainty, since the limits are fixed and known at design time. 
This represents epistemic uncertainty, as the imprecision arises from incomplete knowledge about a component's exact behavior rather than inherent randomness. 
The occurrence source is the hardware itself, adaptation is managed internally by the robot, and the scope is local and component-level. 
The associated risk is low, given that manufacturers' datasheets provide precise data; the affected quality attributes are safety and reliability, as components violating specifications directly compromise both; failure propagation is isolated rather than cascading; and the ethical consideration is transparency, as disclosing hardware tolerances is fundamental. 
Similarly, the remaining taxonomy can be interpreted in the same way. 
The complete taxonomy is available at~\url{https://github.com/Simula-COMPLEX/uncertax}.

\subsection{Prompting}\label{ssec:prompting}
\approach{} features four prompting methods: \textit{role-based prompting} to establish persona-driven context, \textit{rubric-based prompting} to incorporate human-assigned rankings for qualitative refinement, \textit{few-shot prompting} to provide examples for experience-based guidance, and \textit{ontology-constrained} prompting to guide the model with taxonomy elements. 
Our approach first identifies the type of user interaction (e.g., initial query or refinement) and, based on this, selects the appropriate prompting method. 
It then constructs a structured prompt by populating a predefined template tailored to that interaction type and prompting method. 
For uncertainty queries, a response format is appended to the prompt that instructs the LLM to structure its output according to the 12 taxonomy dimensions and provide concise reasoning for each. The resulting prompt is then sent to the LLM for processing.

\subsection{\approach{} Workflow}\label{ssec:workflow}
\approach{} follows a three-step process to support the uncertainty analysis, as shown in~\Cref{fig_approach}.

\subsubsection{Step 1 - Context Comprehension}
This step aims to provide the LLM with the context of the robotic case for which uncertainty analysis will be performed in subsequent steps. 
The user (a robotic engineer) provides the system requirements, a role for the LLM, an objective, instructions, and restrictions. 
\approach{} supports an \textit{assistant} role, which instructs the LLM to provide exploratory and collaborative reasoning to the user. 
Based on this input, \approach{} constructs a structured \textit{role-based prompt} that explains the purpose of each element and how to use it to comprehend the robotic context. 
This prompt is then sent to the LLM, which returns a summary of its understanding. 
If needed, the user can adjust the objective, instructions, or restrictions to refine the LLM’s contextual understanding before proceeding to Step~2. 

\subsubsection{Step 2 - Initial Query}
In this step, the user provides an uncertainty-related question to explore in the context of the given robots.
\approach{} automatically constructs a prompt that asks the LLM to respond to the question. The prompt includes a structured response format aligned with the 12 taxonomy dimensions. 
The LLM is instructed to categorize its response along each of these dimensions and to provide concise reasoning that justifies each assigned category. 
The resulting response is then presented to the user, who reviews it and decides whether refinement is needed (and, if so, proceeds to Step~3).

\subsubsection{Step 3 - Iterative Refinement}
If the user is not convinced of a particular categorization and considers the response needs improvement, this step enables prompt-based iterative refinement until the desired response is achieved. 
\approach{} offers three refinement options: ranking-based refinement, example-driven refinement, and taxonomy-guided refinement. 
In ranking-based refinement, the user assigns a score from 1 (lowest) to 10 (highest) to each dimension, where higher scores indicate convincing responses and lower scores highlight dimensions that require improvement. 
In example-driven refinement, the user provides example scenarios from their own experience to better illustrate the intended interpretation. 
In taxonomy-guided refinement, the user selects a closely related taxonomy element to help LLM improve its response. 

For each selected refinement option, \approach{} automatically maps it to one of the prompting methods, as indicated by the color-coding in~\Cref{fig_approach}. 
For example, when the user chooses ranking-based refinement, \approach{} applies \textit{rubric-based prompting}, builds a structured prompt that embeds the ranking scores for each uncertainty dimension, and includes refinement instructions.
The prompt is sent to the LLM, which returns an updated structured response categorized using the same taxonomy dimensions as in Step~2. This refinement cycle continues until the user is satisfied with the response, at which point they may return to Step~2 with a new uncertainty question.

\subsection{Implementation}
We implemented \approach{} as a web application with an Express-based backend and a React-based frontend. 
The frontend offers an interactive interface for structured prompting and iterative refinement, along with a chat view that presents prompts and LLM responses in a structured format. 
The backend uses OpenAI client to access LLMs, with Gemini 2.5 Flash as the default model. 
This choice was motivated by two reasons: (i) it was among the best-performing models~\cite{sartaj2025identifying}, and (ii) at development time, it was the first Gemini model to support hybrid reasoning, thinking capabilities, and a 1M-token context window. 
Although our experiments used Gemini, the client is compatible with multiple models, including ChatGPT and Llama. 
The implementation of \approach{} is available at~\url{https://github.com/Simula-COMPLEX/RoboULM}.

\section{Evaluation}\label{sec:tool-eval}
The objective of the evaluation is to assess the usability and effectiveness of \approach{} in assisting robotic industry practitioners with the systematic exploration of uncertainty.  

\subsection{Evaluation Design}
We evaluated the tool using the four robotic use cases (AMR, IDR, CMR, and AV) with 16 practitioners, four per case study. 
We intentionally included practitioners who work with robots in industry and project researchers focused on robotic case studies, ensuring both practical and research perspectives. 
To accommodate availability and location, we conducted both in-person and online sessions. 
Among the 16 participants, eight joined remotely and eight attended in person. 
Each session was structured into three parts: (i) a 15-minute introduction, (ii) a 40-minute tool activity, and (iii) a 5-minute feedback questionnaire.

\subsection{Evaluation Procedure}
Each session began with an introduction in which we presented the study motivation, the uncertainty taxonomy, tool demonstration, and addressed the participants' questions. 
For the tool activity, participants received a URL to access the web interface and a requirements document, with remote participants connecting via Ngrok and onsite participants through the local network. 
On first access, the tool displayed an informed-consent page explaining the study purpose and stating that no personal information would be processed. 
Participants' task was to explore at least five uncertainty-related questions and apply each refinement method at least three times. 
Clarification questions were welcomed throughout.

\subsection{Data Collection}
In the final part, participants completed a Google Forms questionnaire covering their professional background and tool-related aspects such as usefulness, ease of understanding, and improvement suggestions.
This feedback formed the primary basis of our analysis. 
In addition, the tool logged participants’ interactions, including prompt setups, uncertainty questions, and refinement methods, enabling supplementary analysis of input patterns, commonly used refinements, and whether participants completed the activity within the allocated time.

\begin{figure*}
	\subfigure[Participants roles and frequency of uncertainty work]{\includegraphics[width=0.5\textwidth]{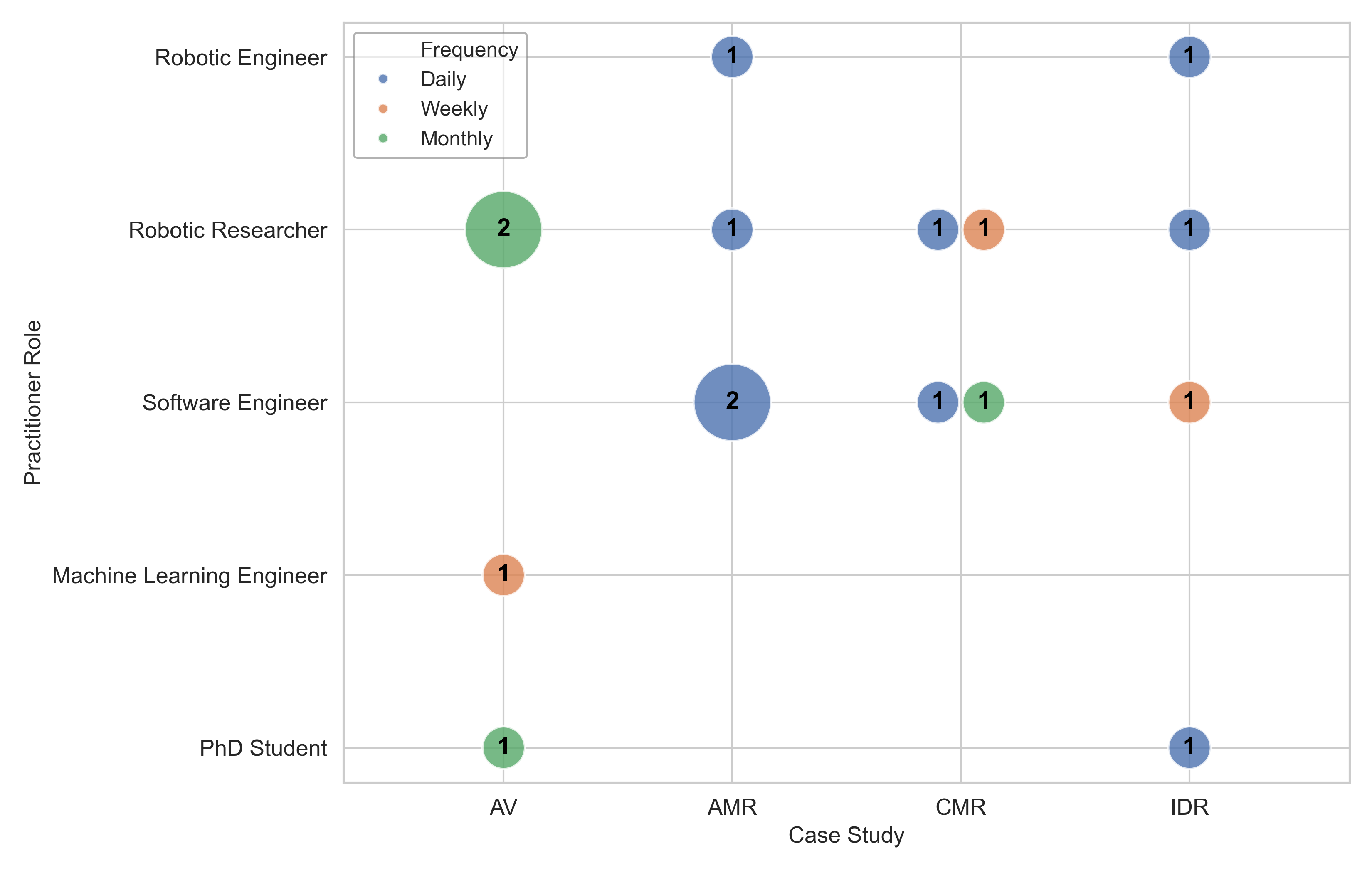}\label{fig:profile}}
    \subfigure[Utility and ease of understanding]{\includegraphics[width=0.5\textwidth,height=5.8cm]{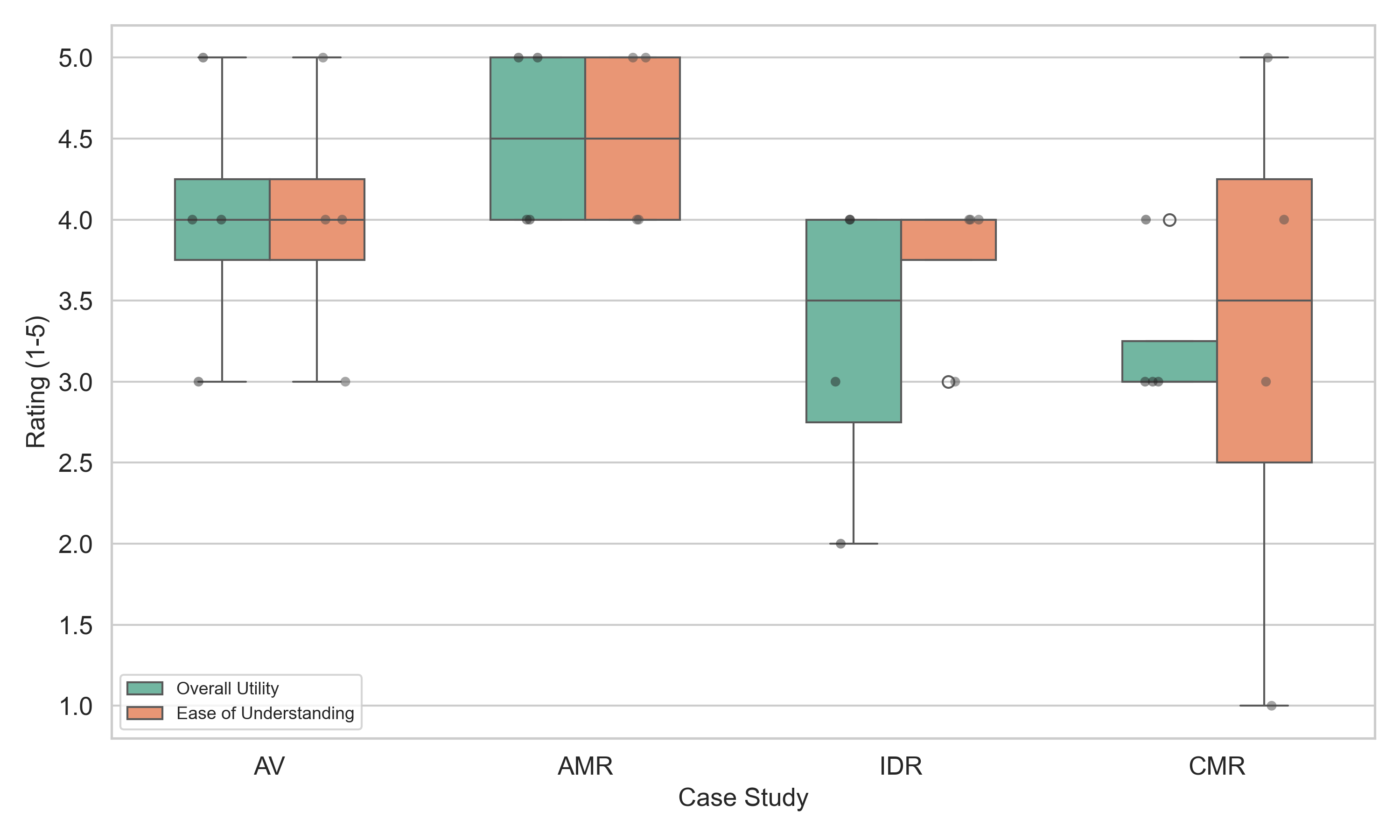}\label{fig:metrics_distribution}}
	\subfigure[Rating of tool features across case studies]{\includegraphics[width=0.5\textwidth]{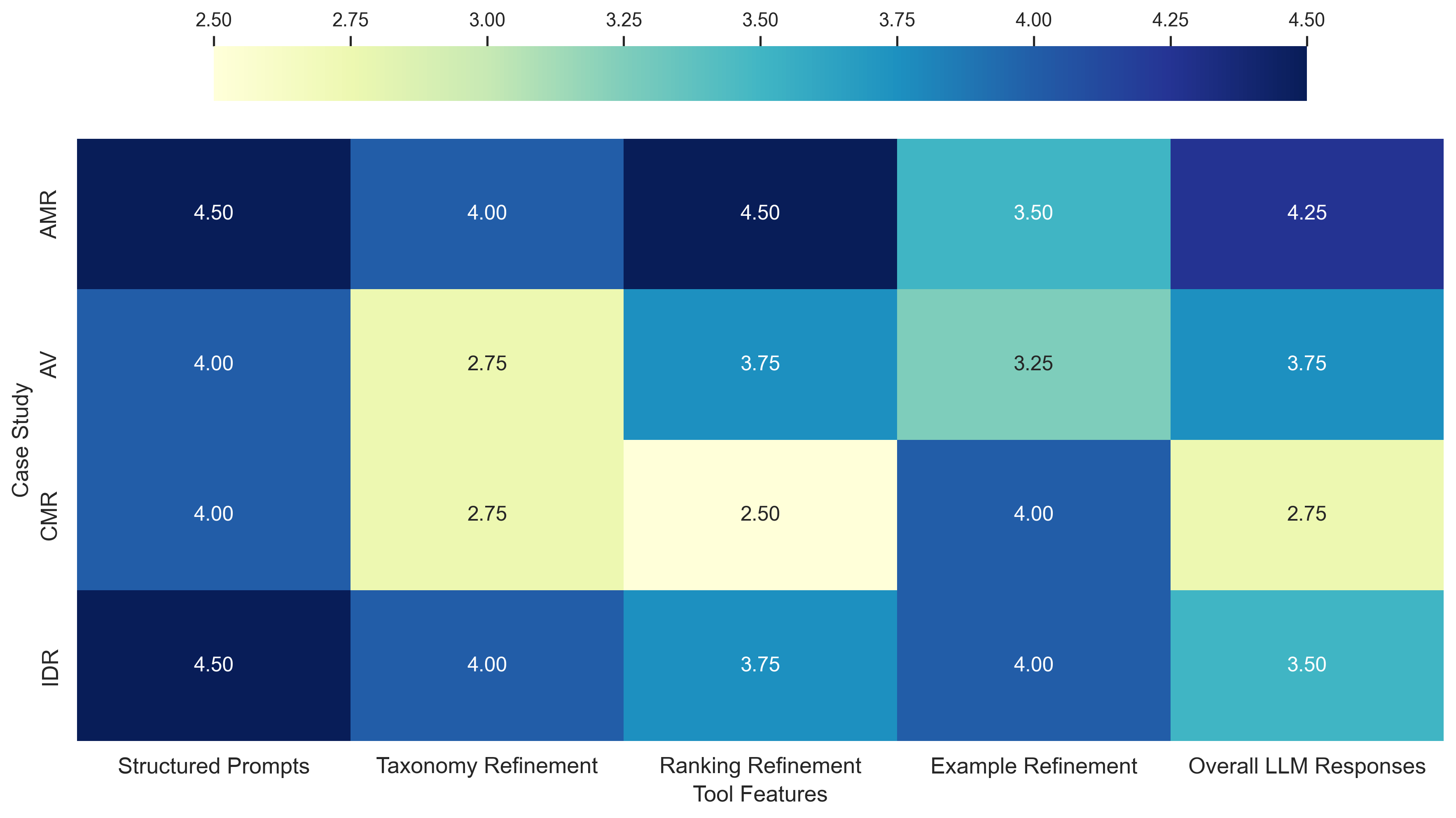}\label{fig:feature_utility}}
	\subfigure[Helpful aspects of the tool]{\includegraphics[width=0.5\textwidth]{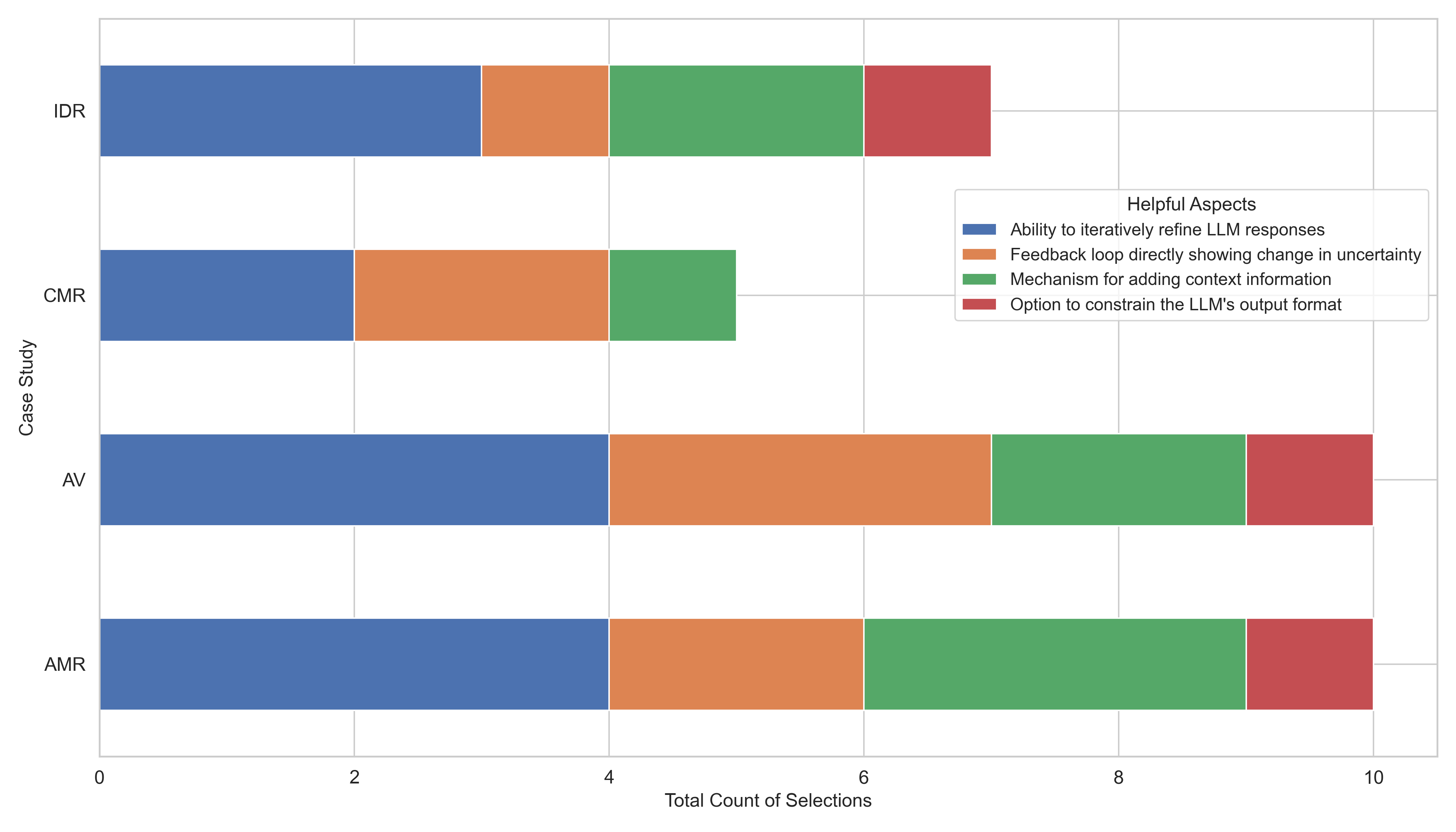}\label{fig:helpful_aspects}}

    \caption{Results of participants' feedback analysis across all four case studies.} 
    \label{fig:allresults}
\end{figure*}

\subsection{Results Discussion}

\subsubsection{Participants Background}
As shown in~\Cref{fig:profile}, each case study included participants with diverse roles and varying frequencies of work with robotic uncertainty. 
Most Robotic Researchers and Software Engineers reported daily or weekly interaction with uncertainty, while all Robotic Engineers in AMR and IDR cases indicated daily engagement. 
Moreover, the group included one Machine Learning Engineer with weekly uncertainty work and two PhD students, one working daily and the other monthly. 
Overall, half reported daily uncertainty work. 
Since the HEU project began in January 2024, most participants had over 2 years of experience with robots. 
Industry participants had hands-on experience throughout the development lifecycle. 
These findings indicate substantial practical/research expertise in managing uncertainty.

\subsubsection{Tool Usability}
In all four cases, \approach{} was generally perceived as useful and understandable, with some variations between case studies (\Cref{fig:metrics_distribution}). 
AMR received the highest ratings (median 4.5/5 for both utility and understanding), followed by AV (median 4.0). 
IDR showed mixed results (utility 3.5, understanding 4.0), while the CMR ratings were widely distributed (understanding median 3.5, overall ratings 1–5). 
Notably, only one participant found it hard to understand, while several others rated the tool above 4.0, highlighting a significant variation in user experience.

\begin{table*}[!t]
\centering
\caption{Descriptive statistics of participants' feedback across all case studies}
\label{tab:descriptive_stats}
\begin{tabular}{lccccc}
\toprule
 & Mean & Median & Mode & Standard Deviation & Top-Two-Box (\%) \\
\midrule
Overall Utility & 3.75 & 4.00 & 4 & 0.86 & 62.50 \\
Ease of Understanding & 3.88 & 4.00 & 4 & 1.02 & 75.00 \\
Structured Prompts & 4.25 & 4.00 & 4 & 0.86 & 87.50 \\
Taxonomy Refinement & 3.38 & 3.00 & 3 & 1.09 & 43.75 \\
Ranking Refinement & 3.62 & 4.00 & 4 & 1.20 & 62.50 \\
Example Refinement & 3.69 & 4.00 & 3 & 0.87 & 56.25 \\
Overall LLM Responses & 3.56 & 3.50 & 3 & 1.26 & 50.00 \\
\bottomrule
\end{tabular}
\end{table*}

\subsubsection{Feature Effectiveness}
Feature evaluation results (\Cref{fig:feature_utility}) show structured prompting as the most consistently valued feature in all cases, with an average 4.5 rating in two cases. 
Among refinement methods, taxonomy refinement was highly rated by AMR and IDR, ranking-based by AMR, and example-driven refinement received strong ratings from IDR and CMR. 
LLM responses received comparatively lower ratings in all cases except AMR due to generating new rather than improved responses. 
Taxonomy-guided refinement was also less favored by CMR and AV participants due to two main factors: the LLM occasionally focused on explaining the taxonomy rather than refining responses, and participants found it challenging to identify the most relevant taxonomy elements to apply.

\subsubsection{Helpful Aspects}
As shown in~\Cref{fig:helpful_aspects}, the most helpful aspect of the tool in all cases was\textit{iterative refinement of LLM responses} (Step~3 of \approach{}). 
The second-most valued were the \textit{mechanism for incorporating contextual information} (Step~1 of \approach{}) and the \textit{feedback loop reflecting changes in uncertainty} (Step~3 of \approach{}). 
Restricting LLM output was the least favored aspect. 
Participants' suggestions point to more flexible interactions, e.g., presenting a concise list of uncertainties for users to select which ones to classify.

\subsubsection{Overall Findings}
\Cref{tab:descriptive_stats} summarizes participants’ evaluations of key features. 
Among all, structured prompting was the most preferred feature, with a mean of 4.25, the lowest standard deviation, and a Top-Two-Box (T2B) score of 87.5\%, indicating that nearly all practitioners rated it as ``Very Useful'' or ``Essential.''
Similarly, the high median, mode, and T2B score (75\%) for ease of understanding suggest that practitioners found the tool intuitive to use, regardless of their technical background. 
Among the refinement methods, ranking-based and example-driven refinement received favorable medians of 4 with some variability; taxonomy-guided and LLM responses scored lower with higher variability due to unpredictable LLM behavior during the process. 
Overall, the tool was perceived as valuable for structured prompting and refinement, suggesting that \approach{} successfully addresses practitioners' needs for systematic uncertainty analysis.

\subsubsection{Analysis of Interaction Logs}
Analysis of tool interaction logs revealed that most participants completed all tasks, some exploring additional uncertainty questions and refinement methods. 
Only two explored a single uncertainty question, but experimented with multiple refinement methods. 
This could be due to time constraints or task misunderstanding; however, neither participant raised time concerns. 
Regarding frequently asked uncertainty questions, participants covered uncertainty sources (environment, hardware, software, human), impacts, failures, mitigations, types, and scenario-based queries. 
Among the refinement methods, ranking-based refinement was used most frequently, likely because it was perceived as straightforward.
Taxonomy-guided refinement was the second most used; however, it was more demanding as participants needed some prior familiarity with the taxonomy. 
Example-driven refinement was the least used method, as participants often struggled to recall relevant examples. 
However, when examples were provided, the refined responses were positively received.

\begin{resbox}[colbacktitle=gray,left=0.5pt,right=2.5pt,top=3.5pt,bottom=3.5pt]{Insights}
\begin{itemize}
    \item Design structured prompts for more consistent, high-quality uncertainty analysis.
    \item Start with ranking-based refinement, then refine with real-world examples. 
    \item Study \taxonomy{} beforehand to effectively guide LLM reasoning. 
\end{itemize}
\end{resbox}

\subsection{Threats to Validity}
A potential threat to \textit{internal validity} arises from the limited session duration and the technical complexity of the tool and uncertainty taxonomy. 
To mitigate, each session included a 15-minute briefing and live demonstration. 
Participants' solid understanding of uncertainty from their daily work (as demonstrated in results) further reduces this threat.

A possible threat to \textit{construct validity} may arise due to social-desirability bias. 
To address this, the feedback questionnaire was kept fully anonymous, and participants were informed that there was no expectation or pressure to provide any particular form of feedback. 

Regarding \textit{external validity}, the generalizability of our results may be limited by the sample size of 16 participants. 
However, previous work suggests that samples of 10±2 participants suffice for usability evaluations~\cite{schmettow2012sample}, and the diversity of industrial cases and participant roles further strengthens the findings.

Finally, a potential threat to \textit{conclusion validity} can arise from drawing conclusions based on a sample of 16 participants. 
To mitigate this, we complemented visual representations with descriptive statistics to ensure a robust basis for interpretation.

\section{Related Work}\label{sec:relatedworks} 
Different studies target uncertainty categorization for adaptive systems and robots~\cite{Harriet24,Honig18,ramirez2012taxonomy}. 
Failures in domestic robotics are explored in~\cite{Harriet24} together with a taxonomy and assessment of the failure impact using the Service Robot Acceptance Model~\cite{FUENTESMORALEDA2020}. 
Ramirez \textit{et al.}~\cite{ramirez2012taxonomy} proposed a taxonomy of potential sources of uncertainty in autonomous systems. 
A taxonomy is proposed in~\cite{Honig18} for robot failures and uncertainty caused by hardware, software, or interactions with the environment. 
While most taxonomies are domain or application-specific, the common denominator is a partitioning of uncertainty features across the robot's lifespan, including design, implementation, and operation phases.

Similarly, effort has been devoted to analyzing and mitigating uncertainties in autonomous robots~\cite{Lafage2025,Wang2025uncertainty,baek2023safety,Icinco25}. Lafage \textit{et al.}~\cite{Lafage2025} proposed an AI-enabled framework that identifies best practices and procedures for uncertainty quantification. 
A framework is proposed in~\cite{baek2023safety} for uncertainty propagation and quantification to assess the safety of robots at runtime. 
Similarly, a digital twin framework is proposed in \cite{Icinco25} for uncertainty quantification and mitigation in autonomous robots.   
Compared to the aforementioned studies, \approach{} offers a customizable, LLM-powered, and human-in-the-loop framework for uncertainty analysis that relies on an uncertainty taxonomy and advanced prompting techniques.

\section{Conclusion}\label{sec:conclusion}
We presented \approach{}, an LLM-based approach and tool to support robotic practitioners in systematically analyzing uncertainties at design time using requirements, alongside an uncertainty taxonomy (\taxonomy{}) for SARs. 
\approach{} incorporates advanced prompting techniques and iterative response refinement methods: \textit{ranking-based}, \textit{taxonomy-guided}, and \textit{example-driven} refinement. 
\approach{}'s evaluation with participants across four robotic industries demonstrated its practicality for systematic uncertainty analysis of SARs.

This article presented a first working prototype of \approach{} together with a usability study. 
In the future, we plan to incorporate participant-suggested features and evaluate them with practitioners from additional robotic cases across multiple LLMs.

\section*{Acknowledgments}
This work is funded by the RoboSAPIENS project under the EU Horizon Europe program (Grant No. 101133807).

\bibliographystyle{unsrt}  
\bibliography{refs}  

\end{document}